\documentclass[sigconf]{acmart}
\usepackage{algorithm}

\usepackage{amsmath}
\usepackage{algpseudocode}
\usepackage{algorithm}
\usepackage{multirow}
\AtBeginDocument{%
  \providecommand\BibTeX{{%
    \normalfont B\kern-0.5em{\scshape i\kern-0.25em b}\kern-0.8em\TeX}}}

\setcopyright{acmcopyright}
\copyrightyear{2022}
\acmYear{2022}
\acmDOI{XXXXXXX.XXXXXXX}

\acmJournal{POMACS}
\acmVolume{37}
\acmNumber{4}
\acmArticle{111}
\acmMonth{8}

\begin{document}

\title{Natural Language Sentence Generation from API Specifications}


\author{Siyu Huo}
\orcid{}
\affiliation{%
  \institution{IBM Research}
  \streetaddress{}
  \city{}
  \state{}
  \country{USA}
  \postcode{}
}
\email{siyu.huo@ibm.com}

\author{Kushal Mukherjee}
\affiliation{%
  \institution{IBM Research}
  \streetaddress{}
  \city{}
  \country{India}}
\email{kushmukh@in.ibm.com}

\author{Jayachandu Bandlamudi}
\affiliation{%
  \institution{IBM Research}
  \streetaddress{}
  \city{}
  \country{India}}
\email{jay_bandlamudi@in.ibm.com}

\author{Vatche Isahagian}
\affiliation{%
  \institution{IBM Research}
  \city{}
  \country{USA}}
\email{vatchei@ibm.com}

\author{Vinod Muthusamy}
\affiliation{%
  \institution{IBM Research}
  \city{}
  \country{USA}}
\email{vmuthus@us.ibm.com}

\author{Yara Rizk}
\affiliation{%
  \institution{IBM Research}
  \city{}
  \country{USA}}
\email{yara.rizk@ibm.com}

\renewcommand{\shortauthors}{Huo, et al.}

\begin{abstract}
APIs are everywhere; they provide access to automation solutions that could help businesses automate some of their tasks. Unfortunately, they may not be accessible to the business users who need them but are not equipped with the necessary technical skills to leverage them. Wrapping these APIs with chatbot capabilities is one solution to make these automation solutions interactive. In this work, we propose a system to generate sentences to train intent recognition models, a crucial component within chatbots to understand natural language utterances from users. Evaluation of our approach based on deep learning models showed promising and inspiring results, and the human-in-the-loop interaction will provide further improvement on the system. 
\end{abstract}




\maketitle

\section{Introduction}
\label{sec:introduction}
% Motivation for interactive automation
In the age of the fourth industrial revolution, a digital transformation journey is critical to the survival of many companies. Primary sectors have adopted automation much sooner than tertiary sectors (like the services industry) \cite{laurent2015intelligent}. One of their main obstacles has been the lack of accessible automation, especially given their lack of technical expertise in the field of software development and automation. Interactive automation through natural language can help achieve more widespread adoption of software automation services, making the digital transformation accessible to a wider range of businesses.

% Chatbots
Conversational systems or chatbots quickly gained popularity in service industries as one of the first forms of automation adopted in the industry \cite{behera2021cognitive}. Despite many of the shortcomings of such systems, especially in their early days, their biggest strength was attempting to speak the same language as humans. Expanding the scope of such systems to automate complex tasks beyond information retrieval requires the integration of conversational systems with automation services.

% APIs
Automation services are generally exposed as API (Application Programming Interface) endpoints that abide by a standard and simplify their integration with other solutions. However, these programs may not necessarily understand or speak the language of humans. To make them truly accessible to non-tech-savvy business users, these automation solutions should be given natural language understanding (NLU) and responding capabilities. Conversational interfaces require time to build and expertise in machine learning and natural language processing. Unfortunately, the developers of these APIs may not have the right background knowledge to create this NLU layer or may not be aware that their APIs would be used in a conversational setting. Therefore, we need an approach to automatically create this layer by leveraging existing information about the APIs (e.g., from their specifications). 

% APIs + chatbots
There have been multiple efforts in the literature to enable conversational interfaces on top of APIs such as \cite{vaziri2017generating} who create a rough chatbot from swagger specifications and improve it by having a power user interact with the chatbot. This effort, and others like it, suffer from a somewhat unnatural way of interacting with the chatbot due to the limitations of the intent recognition models that may not be trained on enough sentences to generalize well. Augmenting such approaches with mechanisms to generate sentences could prove useful, albeit difficult. To date, AI technology is not flawless and may produce irrelevant or grammatically incorrect or ambiguous sentences. It is important to include the human in the loop to provide feedback and ensure higher quality sentences. 

% Our approach and results
In this work, we present a framework that uses deep learning methods to generate sentences from OpenAPI\footnote{https://spec.openapis.org/oas/latest.html} specifications with a human-in-the-loop option and the following goals in mind: 
\begin{itemize}
    \item Extensibility: plugin any NL generation techniques,
    \item Composability: combine (ensemble) one or more NL generation techniques,
    \item Customizability: configure the NL generation pipeline,
%    \item Customizability: users should be able to use any NL generation technique,
%    \item Extensibility: ability to change and remove techniques,
%    \item Scalability: combine (ensembles) one or more NL generation techniques,
    \item Ease of Use: easily packaged and configured by an admin user.
\end{itemize}

The resulting classification models can be used in standalone chatbots or in multi-agent chatbots \cite{rizk2020conversational}. Our extensive experimental evaluation using state of the art deep learning models on publicly available datasets highlight the efficacy of our technique. Our sentence generation method is currently deployed as part of IBM's Watson Orchestrate product.

% Paper Outline
The remainder of this paper is organized as follows. In Section \ref{sec:relatedwork} we present recent related work and highlight the differences of our approach. In Section \ref{sec:architecture}, we present a detailed architecture of our framework. In Section \ref{sec:Deployed System}, we show a demo of our deployed system in a real application. We evaluate our approach in Section \ref{sec:eval} and conclude with final remarks in Section \ref{sec:conc}.

\section{Related Work} \label{sec:relatedwork}
Next, we briefly survey existing work on generating natural language interfaces to various applications before focusing on natural language generation techniques for the purpose of intent classification. 

\subsection{Natural Language Interfaces}
% for databases
Automatic generation of natural language interfaces have been investigated for many applications. Querying databases through natural language is one such application; methods to translate natural language phrases to formal queries vary from semantically parsing the phrases before mapping them \cite{mittal2021bootstrapping, han2020bootstrapping} to deep learning sequence to sequence models \cite{hosu2018natural}. 

% for APIs\
Natural language interfaces have also been automatically generated from OpenAPI specifications to wrap APIs. Vaziri et al. generated a crude chatbot model from Swagger files and engaged a power user to interact with the system for it to improve over time before deploying it to end users \cite{vaziri2017generating}. On the other hand, \citet{babkin2017bootstrapping} adopt an approach that does not include a human in the loop; instead, they augment the training dataset for intent classification training with unsupervised and distantly-supervised approaches. They use distance metrics in sentence embedding spaces to find similar sentences to those in the Swagger file. 

% for command line ?
% Landhäußer, Mathias, Sebastian Weigelt, and Walter F. Tichy. "NLCI: a natural language command interpreter." Automated Software Engineering 24, no. 4 (2017): 839-861.

\subsection{Natural Language Generation}
Intent classifiers are one of the key components of conversational systems \cite{Collinaszy2017}. Commercial products such as IBM Watson Natural Language Classifier and IBM Watson Assistant employ support vector machine (SVM) classifiers. While recent models may be based on neural networks, these SVM models have the advantage of performing well with smaller data sets, requiring less resources to serve the models, and providing a quicker response time \cite{Qi2021BenchmarkingCI}. 

For a new chatbot, a developer may start with a set of fewer than 5 sample utterances per class. Data augmentation via natural language generation (NLG) for handling scarce data works by synthesizing new data from already provided sample utterances, with the goal of improving the performance of the intent classifier.  Basic NLG methods for text typically involve replacing words with synonyms, or changing the word order, as suggested by \cite{Wei2019}. Advanced NLG models facilitate handling limited data by using deep learning methods \cite{Ratford2018}. Language-model-based data augmentation (e.g., GPT-2 based data augmentation) has also been shown to increase the accuracy of various classifiers including SVM, LSTM, and BERT \cite{Ateret2020}. 

\section{Natural Language Generation Pipeline} \label{sec:architecture}
\begin{figure}
    \centering
    \includegraphics[width=\linewidth]{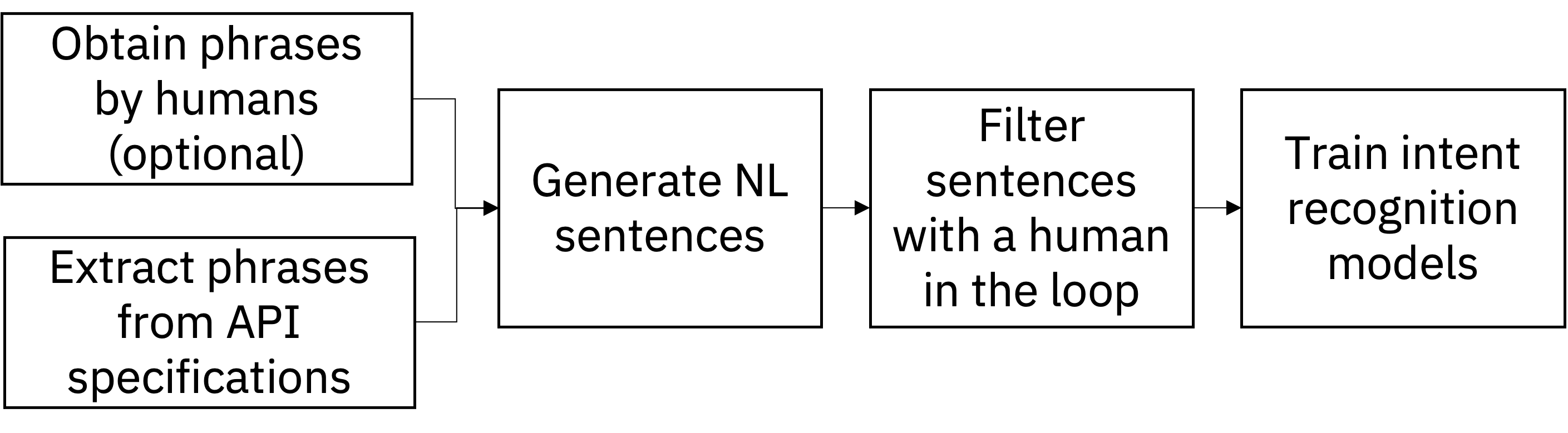}
    \caption{Intent Recognition Model Training Flow from Open\-API Specifications}
    \label{fig:flow}
\end{figure}

\begin{figure}
    \centering
    \includegraphics[width=\linewidth]{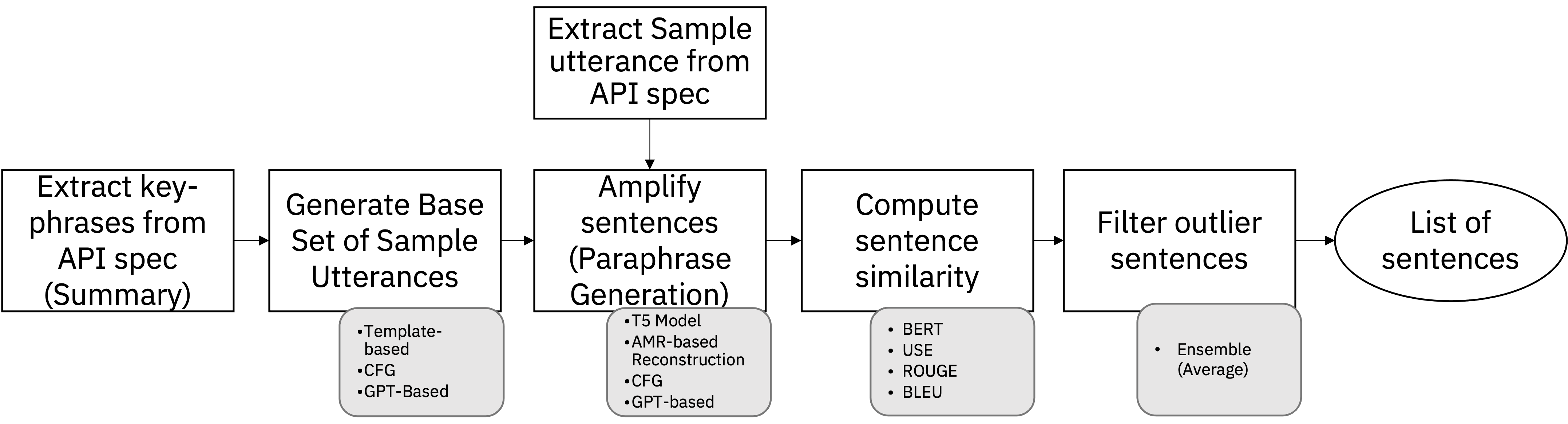}
    \caption{Sentence Generation Flow}
    \label{fig:nlg_flow}
\end{figure}

As stated in Section \ref{sec:introduction}, our goal is to expand the scope of enterprise applications by enabling a conversational interface on top of APIs. As shown in Fig. \ref{fig:flow}, first, we start by extracting phrases from the OpenAPI specification (and possibly augmenting them with human-provided phrases). Second, we generate equivalent sentences using a variety of language models based on the flow described in Fig. \ref{fig:nlg_flow}. Third, since not all the generated sentence may be of high quality, we filter them to eliminate noisy sentences and select a diverse set of sentences (this can be done with a human in the loop or other approaches). Finally, we train an intent recognition model for the chatbot. In the following subsections, we will dive deeper into each one of these blocks. 

\subsection{Extraction of action phrases and initiation of sentences from the API description (1/4)}

Chatbots mimicking human-human interaction or goal-oriented dialogue systems typically converse with humans based on defined tasks \cite{r:dan}. The dialogue systems process customer requests and send queries to the back-end web APIs to retrieve business data and complete certain tasks. For the goal-oriented dialogue systems, the intents from customer requests which the dialogue system can recognize and support are directly related to the capabilities of the back-end system. These intents are created by the dialogue system developer based on the API resources and possible operations defined on those resources. In general, API resources we are interested in include: 1) functional and behavioural servers where the resources can be accessed through their endpoints via HTTP protocols, and 2) the descriptions of each endpoint created by the API server developers. Recent work shows that these intent-associated descriptions of the API server can potentially guide the generation of dialogue flows \cite{Simsek2018MachineRW}. 

Currently, there are a few API description languages such as OpenAPI, WADL\footnote{Web Application Description language}, RAML\footnote{https://raml.org} and API Blueprint\footnote{https://apiblueprint.org}. Most RESTful services are built on top of these lightweight API specifications which allow all consumers of the service (including other developers, customers, or even other services) to understand and interact with the service with minimal implementation logic. For example, smartAPI \cite{Zaveri2017smartAPITA} adopts the OpenAPI specification and allows web API developers to attach semantic metadata to their API descriptions.

\begin{figure}
    \centering
    \includegraphics[width=\linewidth]{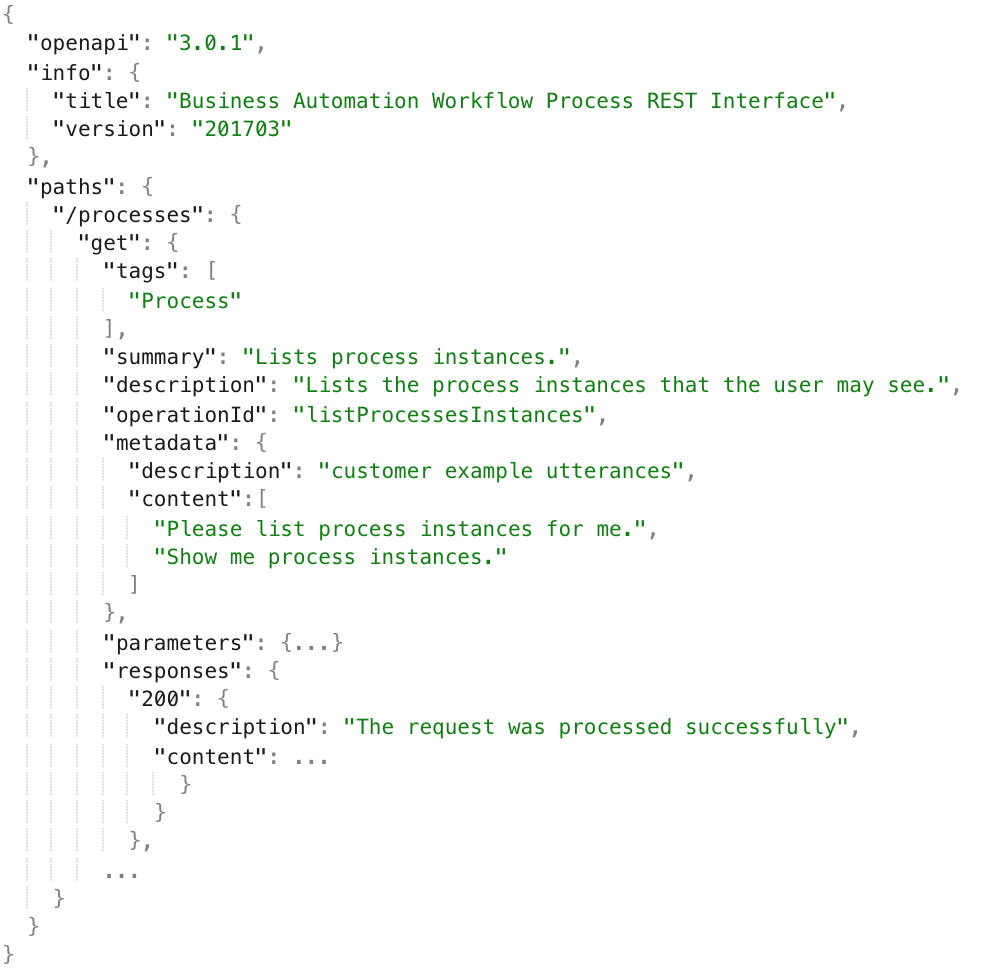}
    \caption{Example OpenAPI specification for an API that lists process instances}
    \label{fig:opanapi}
\end{figure}

Our dialog system is currently based on web APIs which are developed by server developers using the OpenAPI specification. Fig. \ref{fig:opanapi} shows a simplified example of an OpenAPI document about listing process instances in business automation services. However, due to server developers having their own choices and preferences in naming, annotating and describing an API endpoint, it is hard to find a unified method to analyze a given API description. Here, we discuss a few common scenarios of API descriptions and our proposed methods to extract intents from these input descriptions and initiate intent-associated text which is grammatically correct to train the intent classifier. Fortunately, in practical chatbot systems development cycles, the API description schema is known. We provide API developers guidance on how to make API descriptions more amenable to conversational artifact generation methods. 

\subsubsection{Scenario 1: Intent from Endpoint Name}
In an OpenAPI description, the ``path'' section defines intent-related individual endpoints in the API, and the HTTP operations (such as GET, POST) that are supported by these endpoints. Each operation can have some operation objects such as operationId, short summary, long description, tags, parameters, requestBody, response and other metadata. 

The operationId of each operation will be used as the name of the function that handles this request in the generated code, and usually it will contain the action phrase in a certain format which reflects the intent of this function. In order to extract the action phrase in this case, we use regular expressions (RE) to match the format and split word tokens, and use context free grammar (CFG) utilizing the verbs and nouns appearing in the tokens to build the full sentence to train intent classifiers. A similar approach to generate sentences based on API descriptions is discussed in prior art \cite{Simsek2018IntentGF}.

\subsubsection{Scenario 2: Intent from Endpoint Functional Description}
The short summary and long description fields contain one or a few sentences which describe the function and intent associated with the API operation. To utilize these language resources, we first apply semantic parsing (such as abstract meaning representation (AMR) \cite{banarescu2013abstract} or dependency parsing \cite{DependencyParsing}) on these utterances to extract direct object (dobj) components (verb-noun phrases). Then, we use a back-generator (or CFG) to generate new sentences based on the phrase at the prioritized location in the parsing tree (close to the root) without any stop words (such as `is', `are', `can' etc.). These new generated sentences are then delivered to next step in the pipeline. 

\subsubsection{Scenario 3: Intent from Metadata Examples}
The metadata in the OpenAPI description could provide more details about how the dialog system can access the API operation. For example, the service developer may add a custom \texttt{"example utterances"} field to the API operation specification and list a few sample sentences that express an intent to invoke this operation. These sentences can be directly used to train the associated intent classifier. 

\subsection{Sentences Augmentation (2/4)}
According to a survey on data augmentation for natural language processing (NLP) \cite{Feng2021ASO}, there are three main categories of methods to increase the diversity of training data examples for a model without explicitly collecting new data. Among them, rule-based methods focus on the manipulation of word tokens \cite{Wei2019EDAED} or modify semantically annotated components \cite{Sahin2018DataAV} within the sentences. However, these transformations are naive and cannot guarantee correctness of grammar and semantics. Interpretation-based approaches interpolate the inputs and labels of two or more real examples and create mixed-up samples for training and loss minimization, which help alleviate some limitations of neural networks, such as memorization of corrupt labels and sensitivity to adversarial examples \cite{Zhang2018mixupBE}. This approach needs a certain amount of training data and also requires modification of data pre-processing and label encoding steps for the target model; thus, it is not a fit for our task. 

Model-based approaches take advantage of deep learning neural network architectures such as LSTM \cite{Hochreiter1997LongSM} and transformers \cite{Vaswani2017AttentionIA} which pushed the boundaries of NLP technology and enabled the development of new high-performance deep learning models such as ELMO \cite{Peters2018DeepCW}, GPT \cite{Radford2018ImprovingLU}, BERT \cite{Devlin2019BERTPO}, GPT-3 \cite{Brown2020LanguageMA} and T5 \cite{Raffel2020ExploringTL}. These pre-trained models can be customized and fine-tuned for the task of text data argumentation with strong effects on performance since they have achieved high performance on predicting language and generating diverse paraphrases \cite{Gao2020ParaphraseAT,Kumar2019SubmodularOD} of the original samples (which is very relevant to our task). For the sentences argumentation step, we utilize existing deep learning models for paraphrasing original sentences that come from step 1 with the following models.

\subsubsection{Paraphrase with Fine-tuned T5}
There are a few existing paraphrase datasets to study paraphrase detection or generation, where an English text is restated using different phrases while maintaining the same meaning. The Quora\footnote{ https://quoradata.quora.com/First-Quora-Dataset-Release-Question-Pairs} dataset-duplication is composed of 400K pairs of equivalent questions based on real samples from the Quora web site. The PAWS dataset \cite{Zhang2019PAWSPA} provides around 100K well-formed paraphrase and non-paraphrase pairs with high lexical overlap designed for paraphrase detection. Here, we only use its Wikipedia split. The ParaNMT dataset \cite{Wieting2018ParaNMT50MPT} provides 50 million English-English sentence paraphrased pairs, which are generated using neural machine translation to translate the Czech side of a large Czech-English parallel corpus. We tested three T5 models, each fine-tuned with one of the three paraphrase datasets: T5-Quora, T5-PAWS and T5-ParaNMT. Our experiments show that T5-Quora performed better than the other models and is the one we used as the generation model in the pipeline. 

\subsubsection{Few shot learning \& transformation with GPT}
GPT3 is an autoregressive language model that performs strongly in a few-shot setting without any gradient updates or fine-tuning. The few-shot learning is done through a prompt, which is a task-specific natural language input, to interact with the model. In our pipeline, we use GPT-Neo, which is a GPT3-like model, and provide paraphrasing requests in the prompt for each input sentence to generate its paraphrased output.

\subsubsection{Back-translation with AMR}
Back-translation \cite{Sennrich2016ImprovingNM} utilizes seq2seq and language models to translate a sentence into another language and then back into the original language. The newly generated sentences are not all identical to the original but keep the same meaning (most of the time depending on the quality of the translation model), thus improving the diversity of the corpus. 

AMR is a semantic representation language that parses a sentence into a rooted, labeled, directed, acyclic graph which provides a clean and canonical representation of the sentence. We believe that the AMR graph is a better intermediate language than back-translation since it avoids the bias of less standardized language translators and thus conserves the original sentence meaning as a more stable intermediate language.   

\citet{Issa2018AbstractMR} show that the same AMR parsed graph can reflect different expressions about the same meaning and bring in diversity of sentences. SPRING \cite{Bevilacqua2021OneST} is the state-of-the-art AMR parsing and generation framework based on a seq2seq BART model \cite{Lewis2020BARTDS}. To utilize this framework, we first use the SPRING parser to parse an input English sentence into an AMR graph, and then use the SPRING generator to translate the parsed graph back to English with diverse decoding \cite{Fan2018HierarchicalNS} settings on the BART decoder to generate sentences in various forms. 

\subsection{Sentence Ensemble and Selection (3/4)}
Balancing the fidelity and diversity \cite{Kumar2019SubmodularOD} of data augmentation in NLP, the distribution of the generated sentences should neither be too similar nor too different from the original. Otherwise there is a risk of either overfitting intent classifier or underfitting it by training on unrepresentative sentences. Thus, we need to consider both fidelity and diversity of the generated sentences used for data augmentation. 

We collect all the outputs from multiple paraphrasing models as candidate sentences. Since these models work in parallel and do not supervise each other, the output paraphrased sentences may be of low quality or have duplicates. For fidelity filtering, we use the MPC-BERT model \cite{Gu2021MPCBERTAP} to generate an embedding of the paraphrased sentences and the original input, and compute the cosine similarity between them. Then we filter out candidates with similarity values below a threshold $\theta$ from the origin. For diversity selection, we prefer paraphrased sentences which provide more unique n-grams and remove those which provide too few \cite{Yang2020GenerativeDA}. We propose Algorithm (\ref{algo:1}) to ensemble and select the collected results from these multiple models. The output sentences $G_{selected}$ are then delivered to the next step in the pipeline section as input training data.  

\begin{algorithm}
   \caption{Sentence Selection}
    \begin{algorithmic}[1]
        \State \textbf{Input}: Generated sentences $S$ from single/multiple models in step 2, original sentence $s_0$, Target size N. Similarity threshold $\theta$, minimum n-gram increasing threshold $\gamma$.   
        \State \textbf{Output}: Selected sentences for training intent 
        classifier.
        \State Initialization: $G_{selected}$ $\longleftarrow \{\}$ 
        \State Fidelity filtering: $S_{filtered} \longleftarrow \{x|x\in S, Similarity(x, s_0)> \theta \} $ 
        \Repeat: 
        \State $x_{max} = argmax_{x \in S_{filtered}} (CountNgrams( G_{selected} \cup \{x\})  - CountNgrams(G_{selected}))$
        \State $\Delta ngram$ = $CountNgrams( G_{selected} \cup \{x_{max}\})  - CountNgrams(G_{selected})$
        \State Add $x_{max}$ to $G_{selected}$ \textbf{If} $\Delta ngram > \gamma $  
        \State Remove $x_{max}$ from $S_{filtered}$
        \Until {$| G_{selected} |$ = N} or  {$| S_{filtered}|$ = 0}
        \State \Return $G_{selected}$
\end{algorithmic}
\label{algo:1}
\end{algorithm}

\subsection{Interaction with user and train intent classifier(4/4)} 
Recently, there have been studies \cite{Qi2021BenchmarkingCI,Liu2019BenchmarkingNL} showing that IBM Watson Assistant (WA) with its enhanced intent detection algorithm performs equally good or better than existing intent classification applications. In this work, we adopt WA as our chatbot authoring tool to train the intent classifier. Our pipeline evaluation experiments are all conducted using WA. We believe our results to be representative of other intent classification applications.  

In this module, we introduce a human-in-the-loop section through a web UI where our pipeline can interact with users (API developers) for a few tasks. Currently, we mainly present all sentences generated from last step and ask users to pick those appropriate ones for training the WA intent classifier. The final trained WA intent classifier is deployed and serves to classify users' intents based on their input utterances in the dialog system while attempting to invoke an API endpoint conversationally. 

\section{Evaluation} \label{sec:eval}
In this section, we evaluate our pipeline on real-world datasets spanning multiple application domains. Our main purpose in doing so is to establish the feasibility of our proposed solutions by: (1) evaluating the accuracy of natural language intent classifier trained with sentences generated by our pipeline, (2) understanding the effect of the proposed sentence ensemble and selection method in improving intent classification accuracy, and (3) understanding the effect of input sentence diversity and their number on intent classification accuracy.

\subsection{Datasets}
We are not aware of an existing dataset for the end-to-end evaluation of our task. Thus, we created our evaluation settings and assumptions based on publicly available intent classification datasets which have been used to benchmark natural language understanding services \cite{Qi2021BenchmarkingCI}.

\subsubsection{HWU64} 
This dataset provides 64 intents, 21 domains and approximately 25,000 crowdsourced user utterances. The corpus is in the domain of task-oriented conversations between humans and digital assistants \cite{Liu2019BenchmarkingNL}.
 
\subsubsection{Banking77} 
This dataset covers around 13,000 customer service queries in the banking domain labeled with 77 intents. It focuses on fine-grained single-domain intent detection \cite{Casanueva2020EfficientID}.

\subsubsection{Clinic150}
This dataset consists of around 22,000 examples which includes 150 intent classes over 10 domains of task-oriented dialog, such as banking, travel, and home. The dataset also provides out-of-scope examples; however, in this work, we only focus on in-scope examples \cite{Larson2019AnED}. 

These three datasets provide their own training, validation, and testing splits or cross validation splits which we adopted in our experiments.

\subsection{Experimental Setup}
The extraction of action phrases highly depends on a given API document and varies across applications and domains. As mentioned earlier, due to the lack of an end-to-end dataset for our task, we sample a few utterances from the intent classification datasets mentioned above to use as the input customer utterance for either the API endpoint operationID, description, summary or metadata. Since we expect the API developer to only provide a few example utterances, or a single line of description for documentation, we only sample less than 10 samples for each intent from the datasets as input to our pipeline. 
We conduct our experiments to understand the effect of each component in the pipeline and present important factors that affect the intent classification performance. We note that in this experiment, we bypassed the user feedback phase of the pipeline. Thus, the output sentences are not selected by users before training intent classifier. As indicated previously, we believe that having the user in the loop filtering these sentences will only help to improve the overall accuracy of the classifier.

%The extraction of action phrases highly depends on given API document varying in different applications and domains. as mentioned ealier, its particularly difficult to find a standard data resource to evaluate this step, instead we opt to sampling a few  sample a few utterances from the intent classification datasets to mimic input customer example utterances retrieved and reorganized from either API endpoint operationID, description, summary or metadata. Reminds that API developer may only list very few example utterances, or only one line of description for documentation, we only sample less than 10 samples for each intent from datasets as pipeline input. Then We conduct our experiments to exam effects of each component in the pipeline and reveal some important factors below that dominate intent classification performances. It should be mentioned that in experiment we skip the user feedback interaction in the pipeline and thus the output sentences are not selected by users before training intent classifier. 

\subsection{Experimental Results}
\subsubsection{User Input Quality}
We assume that good user input should be diverse enough to cover all aspects of the target intent. For example, for the intent \emph{set alarm} in the home domain from HWU64, a diverse input could be \{\emph{set alarm at 9a.m.; wake me up at 2p.m.} \} versus redundant examples \{\emph{set alarm at 9a.m. ;  Please set the alarm around 2p.m.} \}. We believe that the first example provides a richer diverse corpus which helps to build an intent classifier that can be generalized for more edge cases. To choose diverse samples among sentences, for each intent in the training data, we apply k-means clustering using the sentence encoder USE \cite{Cer2018UniversalSE} as the distance metric to divide the sentences into $n$ different clusters. For each cluster center, we pick the closest sentence to the center as the representative sentence for this cluster (measured by Euclidean distance of sentence embedding). Thus, $n$ clusters will result in $n$ \emph{diverse representative} sentences (one sentence per cluster) which will be used as user input. In contrast, we pick $n$ sentences around the same center of the smallest cluster (the cluster has the fewest number of examples), and we view these sentences as the \emph{narrow representatives} group. As a baseline comparison, we randomly sample $n$ sentences from the same intent to be used as an example of a user's input. We summarize our sampling methods in Algorithm (\ref{algo:2}).

\begin{algorithm}
   \caption{Sample different groups of representatives for one intent from the training set}
    \begin{algorithmic}[1]
        \State \textbf{Input}: training data $D$ for specific intent.    
        \State \textbf{Output}: $Diverse_n$, $Random_n$, $Narrow_n$ groups, each contain $n$ sentences. 
        \State  Initialization: $Diverse$ $\longleftarrow \{\}$, $Random$ $\longleftarrow \{\}$, $Narrow$ $\longleftarrow \{\}$  
        \State $Clusters$, $Centers$  = Kmean\_cluster(embed($D$),n), $|Centers|=n$, $Centers_i \in Clusters_i$ for $i=1...n$ , $ \Sigma_{i\in 1...n} |Clusters_i| $ = $|D|$
        \State Smallest cluster:  $Clusters_{c}$ =$argmin_{c \in 1...n}(|Clusters_c|$) 
        \State $Diverse$ = \{$x_i |x_i$=$argmin_{x \in D}$(distance(x, $Centers_i$), for $i=1...n$ \}   
        \State $Random$ = \{$x_i |x_i = D_{random\_index\_i}$, for $i=1...n$ \} 
        \Repeat
        \State
        $x_{min}$ = $argmin_x$((distance(x, $Centers_{c}$)) 
        
        Add $x_{min}$ to $Narrow$
        
        Remove $x_{min}$ from $D$ for next iteration.
        \Until { $|Narrow|$ = n}
    
        \State \Return $Diverse_n$, $Random_n$, $Narrow_n$

\end{algorithmic}
\label{algo:2}
\end{algorithm}

We use the sentences from each of the three groups mentioned above as the input base sentences of our pipeline, and configure T5-Quora as the sentence generation model for all three cases. We test the final WA performance where each was trained with the enhanced sentences of the pipeline outputs for one group. Results of the classification accuracy on the test data are shown in Table(\ref{table:different inputs}). Note that throughout this and all remaining experiments below, $n$ is the number of input sentences, and the total number of sentences generated using our enhanced pipeline is $5n$.

% As expected, Table(\ref{table:different inputs}) highlights that diversity is crucial in training a good NLU classifier. As the number of diverse sentences increase, the higher the accuracy of the classifier. 

Table(\ref{table:different inputs}) shows that the final intent classification improves significantly when inputs are diverse representative sentences compared to the random case (the former averaged 11.9\% better than the latter) and narrow representatives (better by 29.8\% on average). Notice that when the input sentence number $n$ = 1, the diverse and narrow results are equal in Table(\ref{table:different inputs}) since there is only one cluster. The two methods will pick the same sentence which is the one closest to the cluster center. We discard this case when we compute their average performance difference. We can thus deduce that to improve the overall pipeline performance, it is necessary to collect a diverse set of inputs from the users.  

%SHOULD WE SHOW THIS? I commented it for now ????
% YARA: Thus we propose to use a model to score users input sentence examples (if given at least two examples) and tell them if these sentences are narrow and need to be adjusted for more diverse expresses. This scoring model can be trained with narrow versus diverse samples collected from datasets we used in this paper, and we release our collections in Github-repo\footnote[4]{Link to github repo} for further research. Since this interaction requires extra user efforts and will cause further interaction problems, we leave it for our future work.

\begin{table}[t]
\begin{tabular}{ |p{1.5cm}||p{1.3cm}|p{1.3cm}|p{1.3cm}| p{1.3cm}| }
 \hline
 \multicolumn{5}{|c|}{HWU64} \\
 \hline 
Input type & $n$ = 1 & $n$ = 2 & $n$ = 4 & $n$ = 8 \\
 \hline
Diverse & \textbf{0.366} & \textbf{0.486} & \textbf{0.632} &  \textbf{0.740} \\
Random & 0.174 & 0.332 & 0.541  & 0.658 \\
Narrow & 0.366 & 0.314 & 0.298 &  0.361 \\
 \hline
 
 \multicolumn{5}{|c|}{Bank77} \\
 \hline 
Input type & $n$ = 1 & $n$ = 2 & $n$ = 4 & $n$ = 8 \\
 \hline
Diverse & \textbf{0.420} & \textbf{0.565} & \textbf{0.718} &  \textbf{0.782} \\
Random & 0.330 & 0.384 & 0.589  & 0.717 \\
Narrow & 0.420 & 0.346 & 0.351 &  0.434 \\
 \hline
 
 \multicolumn{5}{|c|}{Clinic150} \\
 \hline 
Input type & $n$ = 1 & $n$ = 2 & $n$ = 4 &  $n$  = 8 \\
 \hline
Diverse & \textbf{0.490} & \textbf{0.679} & \textbf{0.795} &  \textbf{0.862} \\
Random & 0.320 & 0.546 & 0.711  & 0.803 \\
Narrow & 0.490 & 0.420 & 0.461 &  0.587 \\
 \hline
 
\end{tabular}
\caption{Classification accuracy based on input quality using T5-Quora model}
\label{table:different inputs}
\end{table}

% At the same time, we calculate the distance of other sentences with the same intent to its closest representative, and compute average of them as \emph{representing distance} using formula as:

% We assume that the smaller \emph{representing distance} values shows that the intent are more closely represented by  representatives we select, or these representatives can well cover the sentences for the same intent. We plot results of same data point in Table(\ref{table:different inputs})

\begin{table}[t]

\begin{tabular}{ |p{2cm}||p{1.1cm}|p{1.1cm}|p{1.1cm}| p{1.1cm}| }
 \hline
 \multicolumn{5}{|c|}{HWU64} \\
 \hline 
Model type &  $n$  = 1 &  $n$  = 2 &  $n$  = 4 &  $n$  = 8 \\
 \hline
T5-Quora & \textbf{0.366} & 0.486 & 0.632 &  \textbf{0.740} \\
T5-PAWS & 0.257 & 0.407 & 0.605  & 0.705 \\
T5-ParaNMT & 0.237 & 0.431 & 0.604 &  0.716 \\
GPT-Neo & 0.341 & \textbf{0.510}  & \textbf{0.641}   & 0.736 \\
AMR & 0.306 & 0.410 & 0.618 &  0.711 \\

 \hline
 
 \multicolumn{5}{|c|}{Bank77} \\
 \hline 
Model type &  $n$  = 1 &  $n$  = 2 &  $n$  = 4 &  $n$  = 8 \\
 \hline
T5-Quora & 0.420 & \textbf{0.565} & \textbf{0.718} &  \textbf{0.782} \\
T5-PAWS & 0.341 & 0.529 & 0.687  & 0.780 \\
T5-ParaNMT & 0.357 & 0.546 & 0.666 &  0.778 \\
GPT-Neo & \textbf{0.443} & 0.553 & 0.686  & 0.763 \\
AMR & 0.378 & 0.535 & 0.699 &  0.767 \\
 \hline
 
 \multicolumn{5}{|c|}{Clinic150} \\
 \hline 
Model type &  $n$  = 1 &  $n$  = 2 &  $n$  = 4 &  $n$  = 8 \\
 \hline
T5-Quora & 0.490 & \textbf{0.679} & 0.795 &  \textbf{0.862} \\
T5-PAWS & 0.381 & 0.590 & 0.791  & 0.849 \\
T5-ParaNMT & 0.401 & 0.626 & 0.785 &  0.844 \\
GPT-Neo & \textbf{0.502} & 0.678 & 0.796  & 0.856 \\
AMR & 0.411 & 0.619 & \textbf{0.799} &  0.845 \\
 \hline
 
\end{tabular}
\caption{Classification accuracy based on different models }
\label{table:different models}
\end{table}

\subsubsection{Model Capabilities}
In order to choose the appropriate paraphrasing models for various situations, we focus on example utterances obtained from API documents which are \emph{diverse representatives} of the intents since those models performed better than the others. Meanwhile, we use the same model settings for diverse decoding methods, maximum length of output, return sample number, etc. to compare their capabilities.

Based on the same \emph{diverse representatives} for each intent as input base sentences, we configure the pipeline with each candidate model, generate the same number ($5n$) of sentences and train the WA models. We compare the performance of these trained WA models on the testing split of each dataset, as shown in Table(\ref{table:different models}).

Several papers \cite{Larson2019AnED,Campagna2020ZeroShotTL,Zhong2020GroundedAF} suggest that data augmentation becomes less beneficial when applied to out-of-domain data, likely because the distribution of augmented data can substantially differ from the original data. Based on the results in Table(\ref{table:different models}), comparing the performance of three T5 models, we found that T5-Quora performed better than T5-PAWS and T5-ParaNMT. We believe the reason for this is that in task-oriented dialog system, queries in conversations between human and their business consultants (Bank77, Clinic150) or digital assistants (HWU64) are more likely to be expressed as a request or question as in the Quora dataset, rather than statements as in the PAWS(wiki) and ParaNMT datasets. Thus, the T5 model fine-tuned with Quora is closer to the domain of task-oriented dialog, and may be able to generate more in-domain sentences and boost the intent classifier's performance. 

The results also highlight the extensibility of our pipeline to utilize different types of models: T5, GPT, and AMR for sentence generation. 

\begin{table}[t]

\begin{tabular}{ |p{2.2cm}||p{1cm}|p{1cm}|p{1cm}| p{1cm}| }
 \hline
 \multicolumn{5}{|c|}{HWU64} \\
 \hline 
Pipeline config &  $n$  = 1 &  $n$  = 2 &  $n$  = 4 &  $n$  = 8 \\
 \hline
Base & 0.348 & 0.435 & 0.588 & 0.697 \\
T5-Quora   & 0.325 & 0.427 & 0.621 & 0.702 \\
T5-Quora+ Selection & 0.366 & 0.486 & 0.632 & 0.740 \\
Ensemble+ Selection &  \textbf{0.399} & \textbf{0.533} & \textbf{0.682} & \textbf{0.776} \\
 \hline
 
 \multicolumn{5}{|c|}{Bank77} \\
 \hline 
Pipeline config &  $n$  = 1 &  $n$  = 2 &  $n$  = 4 &  $n$  = 8 \\
 \hline 
Base & 0.335 & 0.516 & 0.649 & 0.746 \\
T5-Quora    & 0.371 &	0.532 &	0.706 &	0.785 \\
 T5-Quora+ Selection &  0.420 &	0.565 &	0.718 &	0.782 \\
Ensemble+ Selection &  \textbf{0.482}& 	\textbf{0.599}& 	\textbf{0.731}& 	\textbf{0.787} \\
 \hline
 
 \multicolumn{5}{|c|}{Clinic150} \\
 \hline 
Pipeline config &  $n$  = 1 &  $n$  = 2 &  $n$  = 4 &  $n$  = 8 \\
 \hline
Base & 0.403 &	0.605 & 0.754 & 0.835 \\
T5-Quora    & 0.458 & 0.619 &	0.792 &	0.840 \\
T5-Quora+ Selection & 0.490 & 0.679 & 0.795 & 0.862 \\
Ensemble+ Selection & \textbf{0.535 } &\textbf{0.698} &\textbf{0.821} & \textbf{0.866}  \\
 \hline
 
\end{tabular}
\caption{Performance based on pipeline module configuration}
\label{table:different modules}
\end{table}

\subsubsection{Ensemble and Selection Capabilities}
Our experimental results above suggest that T5-Quora, GPT and AMR perform differently depending on the number of sentences, intent type and dataset. Taking the combined advantage of these models is what motivated our pipeline design choice. Thus, in this section, we evaluate the scalability of our pipeline to combine results from multiple models. 

We first examine the effect of the sentence selection component in Algorithm(\ref{algo:1}) on the single model (T5-Quora) output. Then, we examine the ensemble mechanism by combining the three models' outputs, in addition to using the sentence selection algorithm.
We use the same data (diverse) and model settings as previous experiments. The results from WA intent classification are shown in Table(\ref{table:different modules}). The $Base$ row lists the results when we use the original $n$ input sentences (representatives) to train the classifier without going through the pipeline or any data argumentation steps.

Table(\ref{table:different modules}) shows that by increasingly adding Paraphrasing (avg. +2.2\%), Sentence Selection (avg. +3\%) and multiple-model Ensemble modules (avg. +3.1\%) into the pipeline, the intent classifier obtains higher accuracy (avg. +8.3\%, maximum +14.7\%, minimum +3.1\% for full integration) than models only trained on bare $n$ input sentences ($Base$), and the performance gain is more obvious when the input sentence size $n$ is smaller.  

It should be noted that in a few cases, the classifier performance did not improve, even becoming worse, when trained with data from the pipeline (with T5-PAWS, T5-ParaNMT). We checked the outputs and found that in some cases one model may only create limited variations based on its inputs.  
The following is an example from the Bank77 dataset. Given input \{ \emph{how soon can i get cards?} \}, the model generates \{\emph{how soon will i get my card? how soon am i supposed to get my card?  how soon will i get the ID card?} \}. These sentences do add some new n-gram features to the vocabulary and conserve the original meaning, but still lack diverse expressions in the domain and may cause model overfitting on these redundant expressions. The final classifier failed to detect the sentence \{\emph{how long will it take to deliver in US?}\} from the same intent category (\emph{card\_delivery\_estimate}) in Bank77. This observation suggests that paraphrasing models have limitations in generating sufficiently diverse examples to cover an intent. We suspect the reason is that the sentence generation models were not specifically fine-tuned with related in-domain data. In the particular, for the example above (from the banking domain), there is a lack of knowledge about card replacement processes which require a delivery system instead of just handing over the card by a neighbor or friend. This can be problematic in some enterprises where getting in-domain data to fine-tune these models may not be feasible. In such cases, it is important to leverage a human-in-the-loop approach to augment the extracted sentences and improve the generated sentences' diversity.

\section{Deployed System}
\label{sec:Deployed System}
\begin{figure}[!b]
    \centering
    \includegraphics[width=1.0\linewidth]{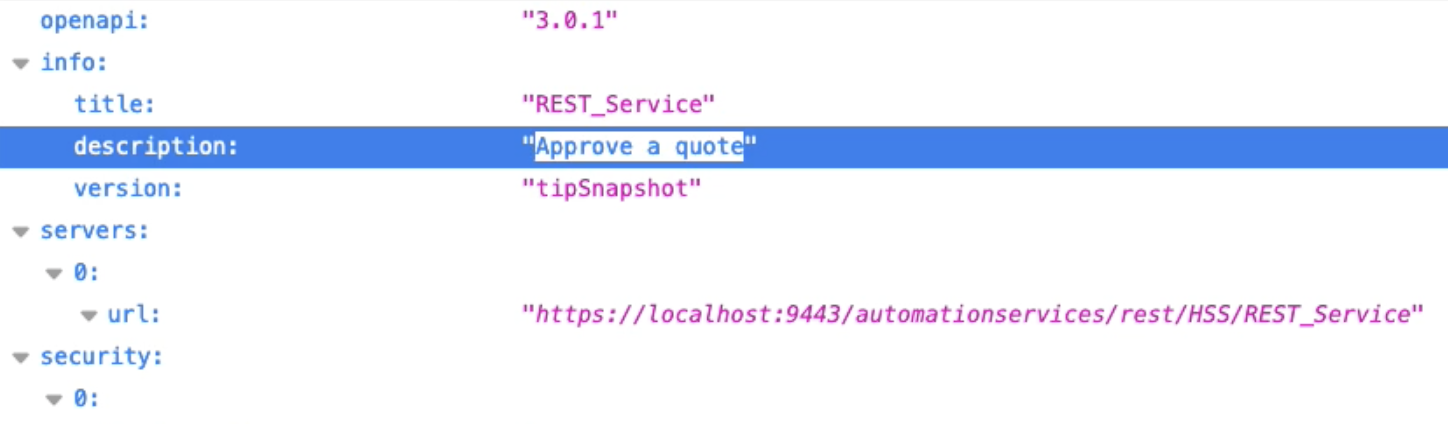}
    \caption{OpenAPI Specification (description)}
    \label{fig:openAPI_bpm}
\end{figure}

\begin{figure}[!b]
    \centering
    \includegraphics[width=1.0\linewidth]{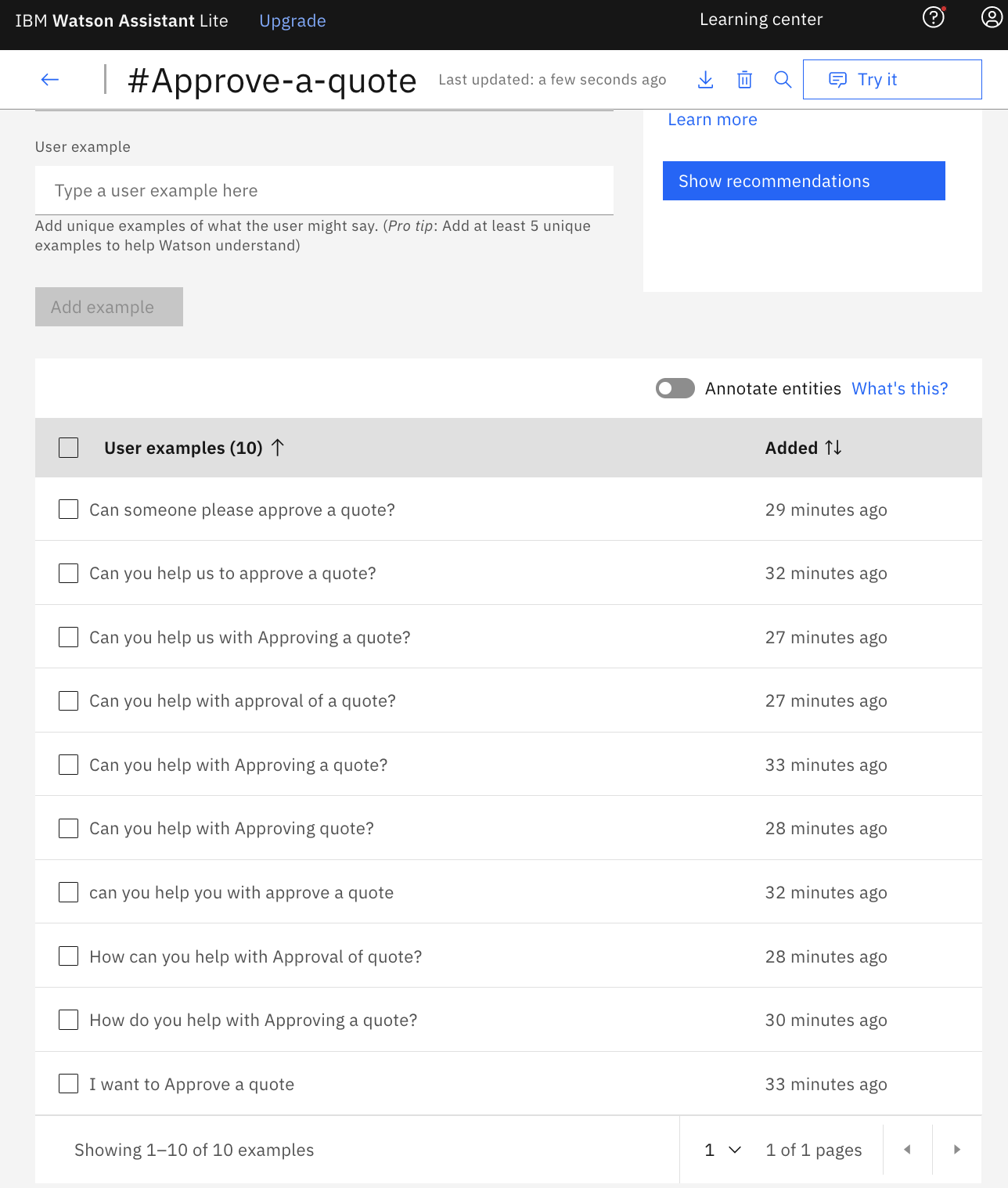}
    \caption{Watson Assistant skill}
    \label{fig:WA}
\end{figure}

\begin{figure*}
    \centering
    \includegraphics[width=\linewidth]{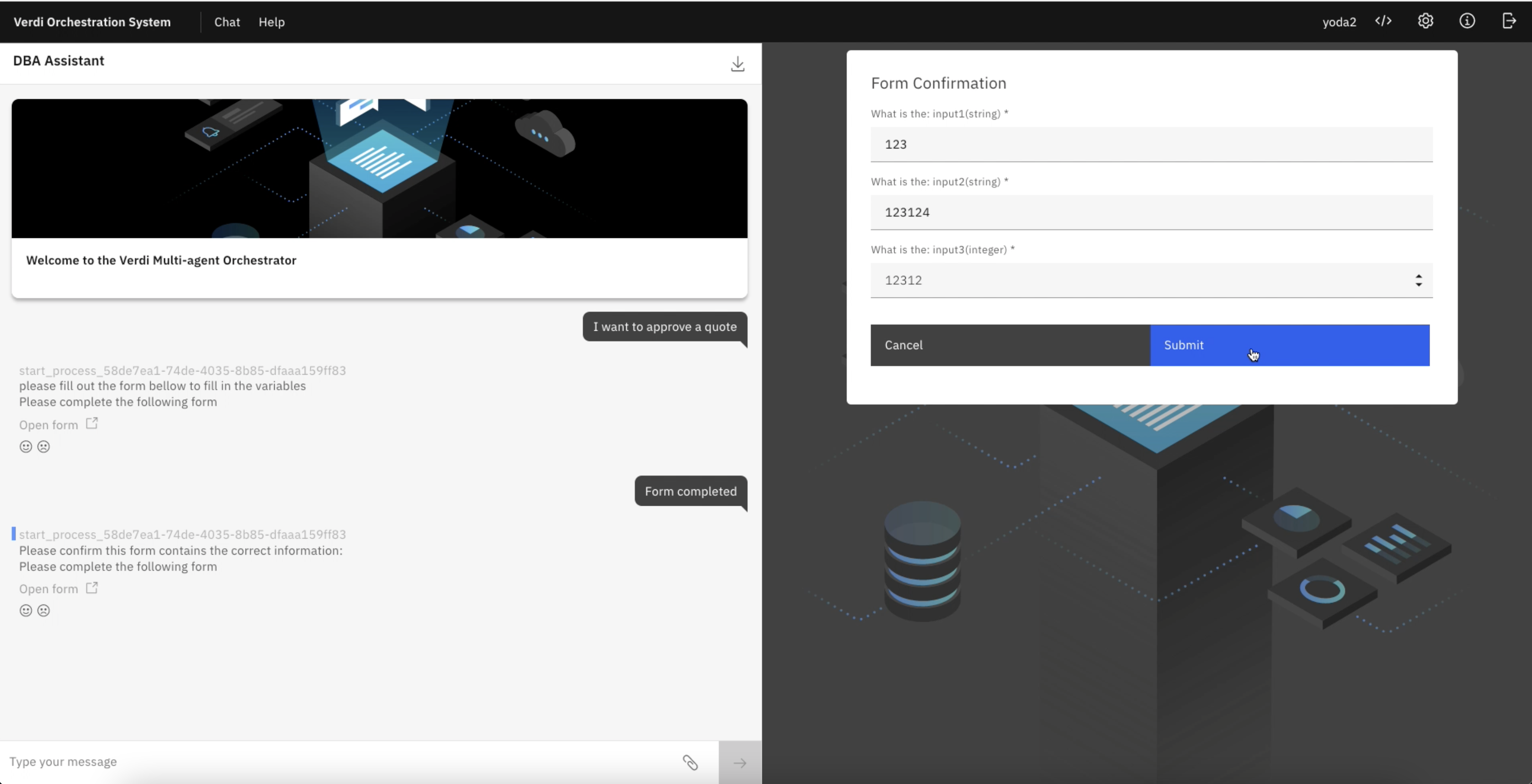}
    \caption{Conversationally invoke the API}
    \label{fig:verdi}
\end{figure*}

The presented approach and a version of our pipeline has been deployed within IBM's Watson Orchestrate product\footnote{https://www.ibm.com/cloud/automation/watson-orchestrate}. It is based on a multi-agent conversational system \cite{rizk2020conversational}. In the recorded demo\footnote{Uploaded as part of supplementary material}, the proposed methodology generates natural language sentences from an OpenAPI specification (see Fig. \ref{fig:openAPI_bpm}) to train an intent recognition model in a chatbot authoring tool like Watson Assistant (WA) (see Fig. \ref{fig:WA}). Once the model is trained, a YAML file defining a conversational agent that includes the generated WA chatbot and the API endpoint is also automatically generated and activated within the multi-agent system. Finally, an end user can conversationally invoke the desired endpoint in the chat interface (see Fig. \ref{fig:verdi}).

\section{Conclusion} \label{sec:conc}
Natural language sentence generation can make chatbot (or at least intent recognition models) creation more accessible to individuals who do not possess the right skills to create them. Our approach proved effective after evaluating it on multiple datasets (accuracy gains were 8.3\% on average using diverse representations) and was deployed within IBM's Watson Orchestrate product. However, the approach does have limitations that we plan on addressing in our future work. This includes taking advantage of domain knowledge when generating sentences, improving the semantic diversity of the generated sentences, and incorporating implicit feedback from users interacting with the chatbot to further augment the training data for intent recognition retraining.

\bibliographystyle{ACM-Reference-Format}
\bibliography{aaai22}

\end{document}